\documentclass[10pt, a4paper, conference , twocolumn]{IEEEtran}

\usepackage{amsmath}
\usepackage{graphicx}
\usepackage{url}

\usepackage{fancyhdr}
\usepackage[a4paper, hmargin=13mm, vmargin={27mm,45mm}, headsep=4mm]{geometry}

\begin{document}
\title{Object tracking in video signals using Compressive Sensing\\
\large Student paper}

\author{
\IEEEauthorblockN{Marijana Kra\v{c}unov}
\IEEEauthorblockA{
University of Montenegro\\
Podgorica, Montenegro \\
Email: kracunov@icloud.com}
\and
\IEEEauthorblockN{Milica Ba\v{s}trica}
\IEEEauthorblockA{
University of Montenegro\\
Podgorica, Montenegro \\
Email: milicabastric@gmail.com}
\and
\IEEEauthorblockN{Jovana Te\v{s}ovi\'{c}}
\IEEEauthorblockA{
University of Montenegro\\
Podgorica, Montenegro \\
Email: jovanatesovic@gmail.com}
}
\maketitle \thispagestyle{fancy} 

\begin{abstract}
Reducing the number of pixels in video signals while maintaining quality needed for recovering the trace of an object using Compressive Sensing is main subject of this work. Quality of frames, from video that contains moving object, are gradually reduced by keeping different number of pixels in each iteration, going from 45\% all the way to 1\%. Using algorithm for tracing object, results were satisfactory and showed mere changes in trajectory graphs, obtained from original and reconstructed videos.
\end{abstract}


\section{Introduction}
\subsection{Compressive sensing}
Thanks to the sparse property, that portrays considerable number of signals that surround us, the engineers are able to use compressive sensing algorithm and so make much better use of electronics today. In order to understand Compressive Sensing [1]-[15] one must first comprehend well famous theorem dating back to 1949. That’s when Claude Shannon, widely known as "the father of information theory", stated: “If a function $f(t)$ contains no frequencies higher than $W$ hertz, it is completely determined by giving its ordinates at a series of points spaced $1/ (2W)$ seconds apart”. Even though this is a fundamental principle in the field of information theory the whole premise of CS lies on avoiding it. Shannon’s theorem states that the resolution of an image is proportional to the number of measurements. Doubling the resolution calls for doubling the measurements. \\
But thanks to the development of CS, the reconstruction of super-resolved images and signals from far fewer measurements than deemed necessary is possible. Not only that it also offers hope for directly acquiring a compressed digital representation of a signal without first sampling that signal.\\
What makes this theorem so applicable is it’s not only restricted to noiseless signals but ,with few or none alternations to algorithm, it also covers the real-world signals which by default include noise.\\
The system inspected here is an under-determined system. An under-determined system is a system in which the measurements taken are less than the number of unknown signal values which results in a system with infinite number of solutions. Fortunately, this problem is solvable if the signal $x$ is compressible.
In order to sense a sparse objects, by taking as few measurements as possible, the best approach is measuring at randomly selected frequencies.
\begin{equation}b=A\bar{x}\end{equation}
$\bar{x} $- discrete signal\\
$\textbf{b}$ – vector of linear measurements formed by taking inner products of $\bar{x}$ with a set of linearly independent vectors $ a_i $,  $i=1,2,3…m$.\\
$\textbf{A}$ – has rows $a_i^T$\\
Encoding is the process of obtaining $\textbf{b}$ from $\bar{x} $ and decoding is the process of recovering $\bar{x}$\\
Presuming that $\textbf{b}$ is derived from a highly sparse signal, meaning it has very few non-zero elements. The best decoding approach is too look for the sparsest signal of all those that take part in producing the measurement $\textbf{b}$
\begin{equation}
\min\lbrace\lVert x \rVert_0:Ax=b\rbrace
\end{equation}
 $\lVert x \rVert_0$-the number of non-zeros in $x$.\\
Due to the complexity of solving this problem by enumeration the "$l_0$ -norm" is replaced by the "$l_1$ -norm"
\begin{equation}
\min\lbrace\lVert x\rVert_1:Ax=b\rbrace
\end{equation}
If we are given a sparse vector $x$ we can write it down as a weighted linear combination.
\begin{equation}
x=\sum_{j=1}^{N}\alpha_j \psi_j
\end{equation}
$\psi_j$ – column vectors \\
$\alpha_j$– Fourier coefficients\\
but only $S$ of these $\alpha_j$ are going to be zero the rest are not going to be zero. $S << N$, $N$ is the number of pixels.
If the signal $\bar{x} $ is sparse enough and with the right matrix $\textbf{A} $ than $\bar{x} $ can solve both (2) and (3) for $b=A\bar{x}$. This is called recoverability.\\
Again, here we are putting an emphasis on actual implementation of CS algorithm in video recordings. Dividing frames from the original video and using CS on every frame independently is in most cases time consuming. Some of the neater solutions can be seen in project regarding this subject [1]. Key idea is to exploit the intra-frame correlation. Using modification of the original Kalman Filter so that the motion prediction is incorporated and as such proceeds to the CS step more beneficially. 
\subsection{Object tracking}
Object tracking is the act of locating a moving object or multiple moving objects over time and reconstructing the trajectory that the objects took [16]-[23].The simplest known method is the block matching technique. While observing the current frame we take a block of pixels into consideration. The idea is to estimate the position by comparing the blocks within a region containing the targeted object with the reference frame [2].The results would be better if the image region occupied by the target in the current frame is visibly different from the other parts of the frame. Problems that occur during object tracking are caused by several factors such as slow frame rate when the object is moving fast, interference caused by a cluttered background etc.[3].\\
   In our study of movement, we got the position, velocity and acceleration of a ball, from a video file recorded with a fixed camera. The detection of this is based on frame subtraction, a common background is subtracted from each frame and then treated. When program works in 2D, we assume that all the movement, velocity and acceleration is contained in a vertical plane perpendicular to the camera. In 3D, the ball is free moving.

\section{CS and object tracking}
Here CS represents a helping hand for managing videos on which OT will be performed. How low can we go on memory storage and still get a satisfying trace that’s actually useful. Problems that are still rising questions are complex background, brightness changes but one of the most challenging are the ones including sudden and/or short ‘disappearance’ of the actual object, i.e. occlusion occurs, in region where installed cameras have no information the cause usually being the object “hiding” behind another. This requires reconstruction of path by assumptions based on its positions before and after vanishing and also on speed and acceleration information.\\
The higher the frequency of frames the better results can be expected. Characterization of an object is an extremely important component for any type of object tracking algorithms. In order to examine all this, we used CS on two different types of signals one that could be described as considerably sparser than the other.
\begin{figure}[h!]
  \includegraphics[width=\linewidth]{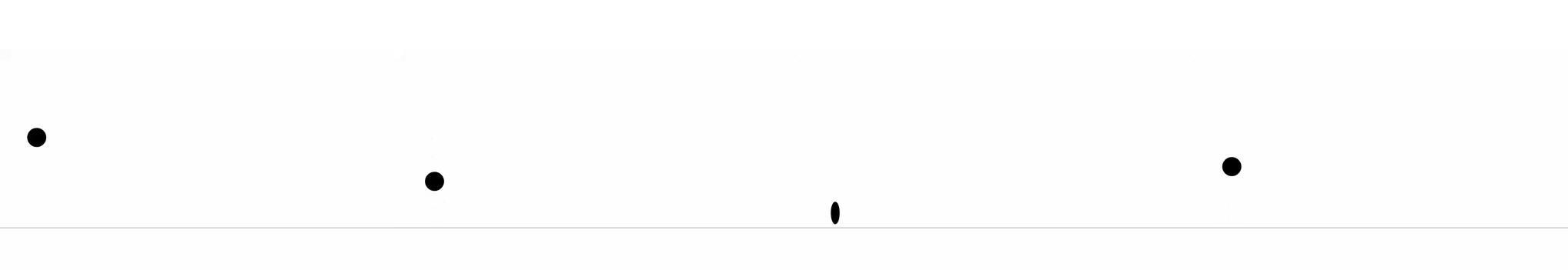}
  \caption{4/75 still frames from original video }
  \label{fig:Video1original}
  \includegraphics[width=\linewidth]{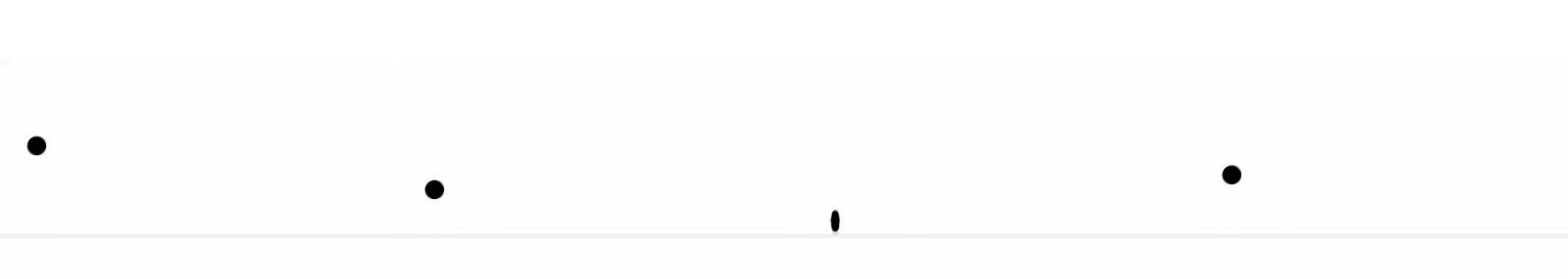}
  \caption{4/75 still frames from reconstructed video (1\%) }
  \label{fig:Video1cs1}
  \includegraphics[width=\linewidth]{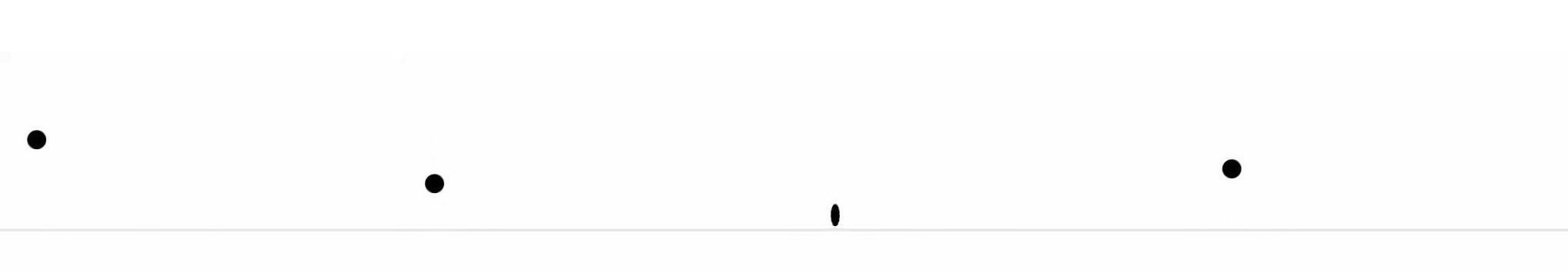}
  \caption{4/75 still frames from reconstructed video (5\%) }
  \label{fig:Video1cs5}
\end{figure}
\section{Experimental Results}
Experiments were conducted using two different videos of a bouncing ball. The Compressive Sensing algorithm used in this experiment requires the video to be converted to grayscale. Then the video is transformed into image frames. Compressive sensing is done on to each frame. The compressed frames are then pieced together to form a new video. Finally the trace of the ball in the video is reconstructed and plotted in 2D.
\subsection*{Experiment I}
\begin{figure}
  \includegraphics[width=\linewidth]{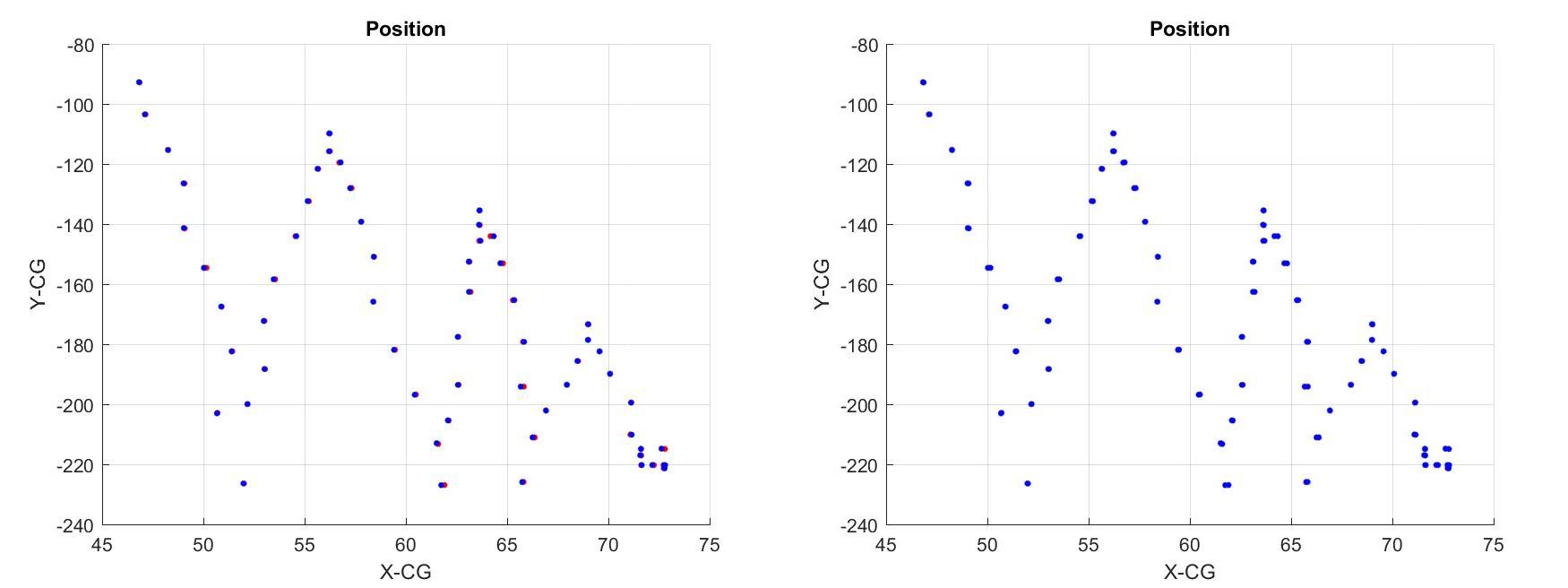}
  \caption{Trajectory of the ball reconstucted from video of 1\%(left) and 5\%(right) of original pixels}
  \label{fig:Video1-CS1/CS5}
\end{figure}

The first video is much more simple, meaning, it is far sparser than the other one. Due to the small number of nonzero coefficients, it serves as a good example of compressive sensing. Figure \ref{fig:Video1original} shows few still frames from the original video.\\
Duration of this video is 2.5 seconds, frame rate is 24.32 frames per second. Unlike the experiment with the other video, which will be discussed later, the background of this video is blank. This is a huge advantage because the background will not cause any interference. As a result of the simplicity of the video, video frames are well reconstructed using 45\%, 30\%, 20\%,10\%,5\% and even 1\% of available pixels. Figure \ref{fig:Video1cs1} and \ref{fig:Video1cs5} present few reconstructed frames from 1\% and 5\% available pixels respectively and figure \ref{fig:Video1-CS1/CS5} presents the difference between the trace of the object when the video is not compressed (red dots) and when only 1\% (left) and 5\% (right) of the pixels were taken (blue dots).\\
As we can see with 5\% we have complete overlapping hence showing results with the higher percentage of conserved pixels is unnecessary.
\begin{figure}[h!]
  \includegraphics[width=\linewidth]{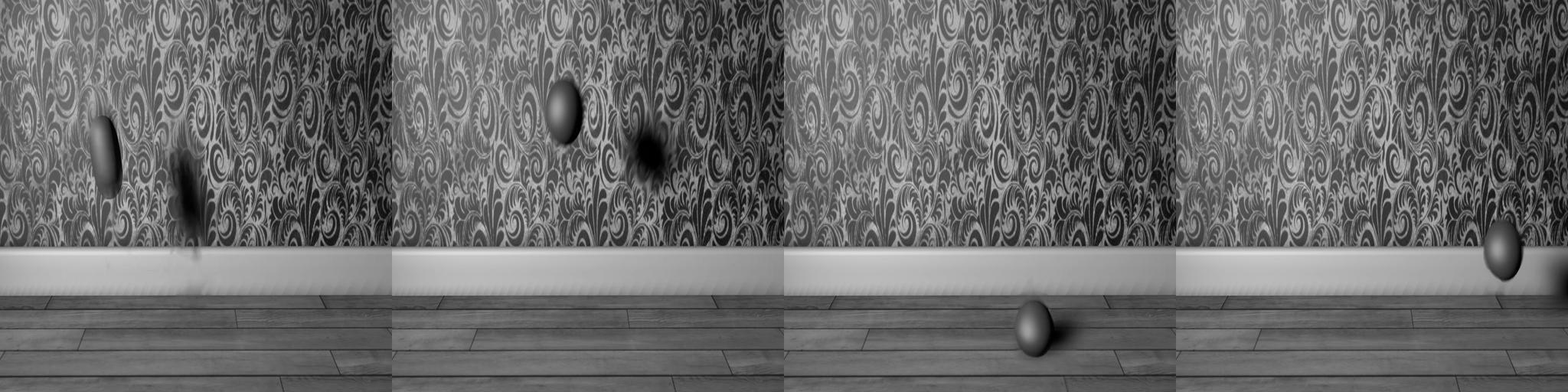}
  \caption{4/75 still frames from original video}
  \label{fig:Video2-ORG}
  \includegraphics[width=\linewidth]{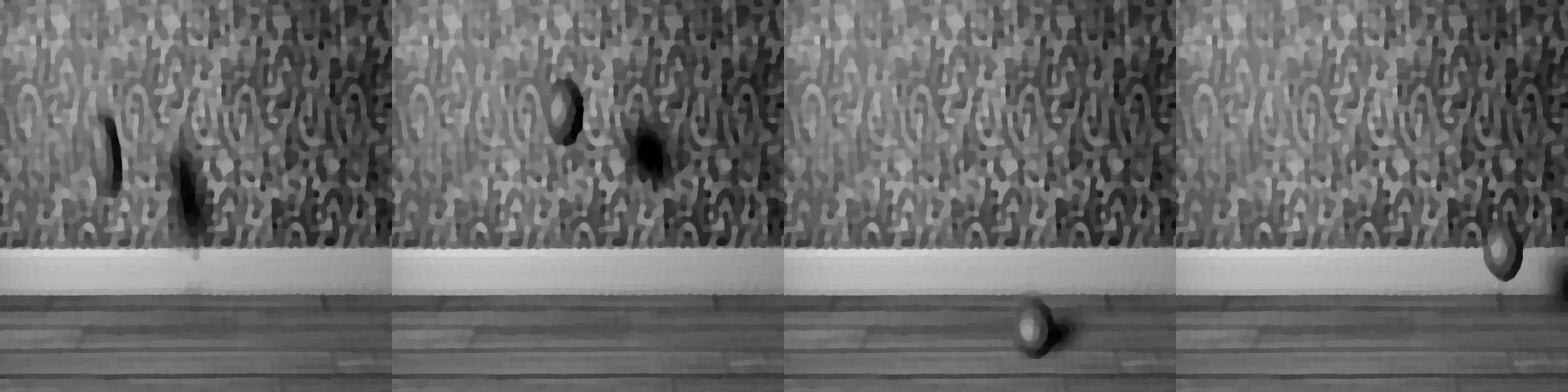}
  \caption{4/75 still frames from reconstructed video (1\%)}
  \label{fig:Video2-CS1}
  \includegraphics[width=\linewidth]{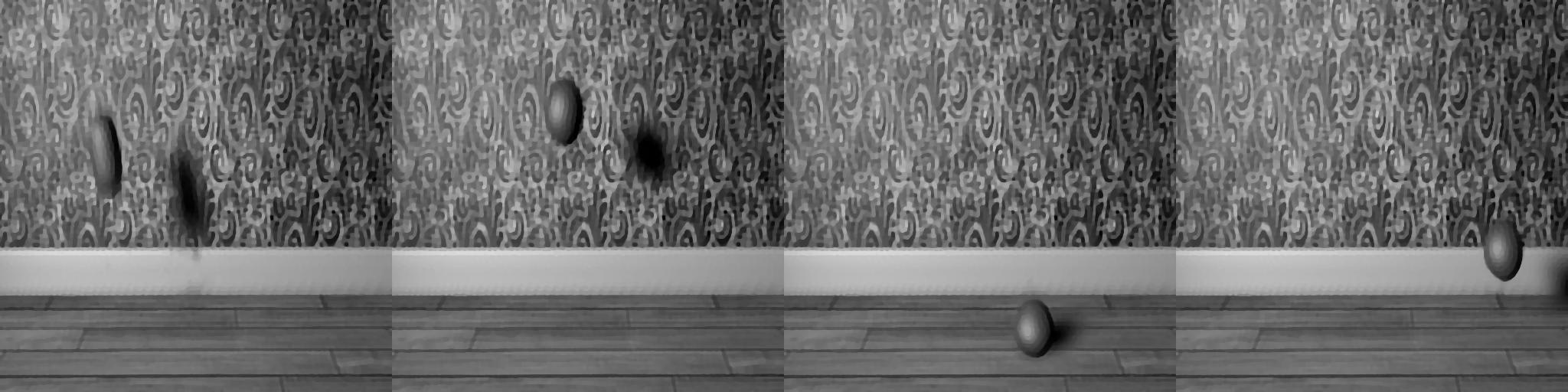}
  \caption{4/75 still frames from reconstructed video (5\%)}
  \label{fig:Video2-CS5}
  \includegraphics[width=\linewidth]{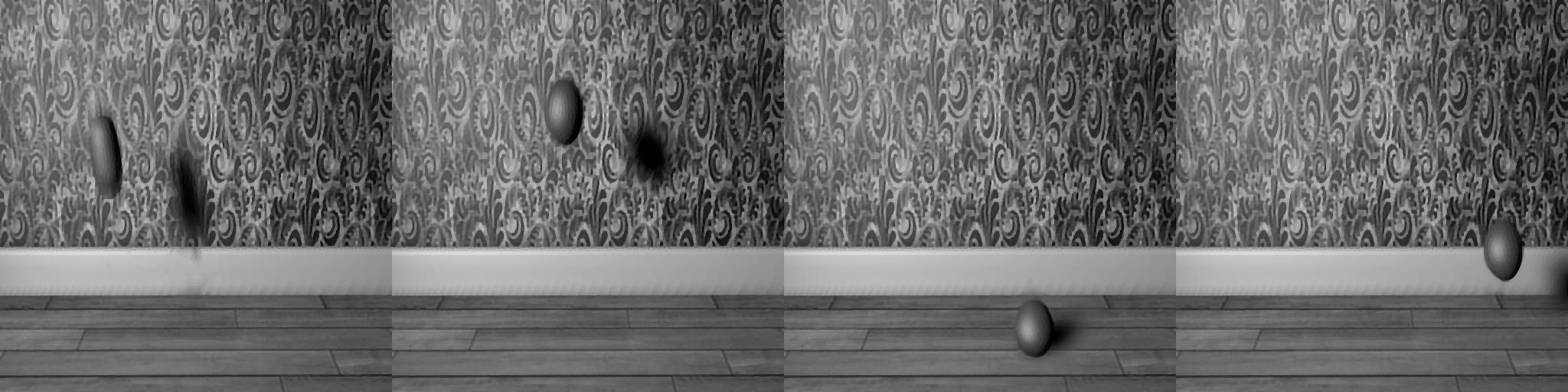}
  \caption{4/75 still frames from reconstructed video (10\%)}
  \label{fig:Video2-CS10}
  \includegraphics[width=\linewidth]{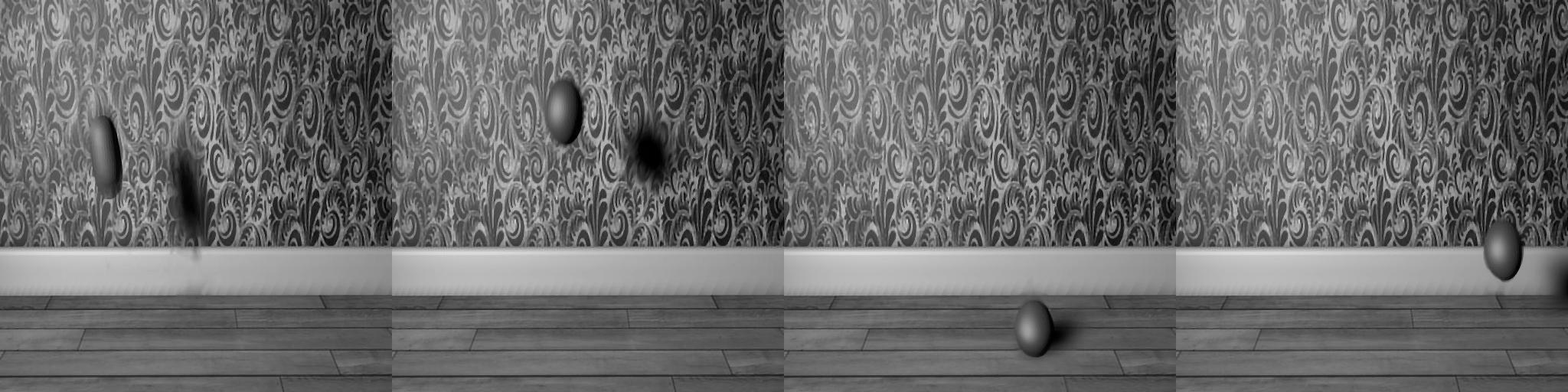}
  \caption{4/75 still frames from reconstructed video (20\%)}
  \label{fig:Video2-CS20}
  \includegraphics[width=\linewidth]{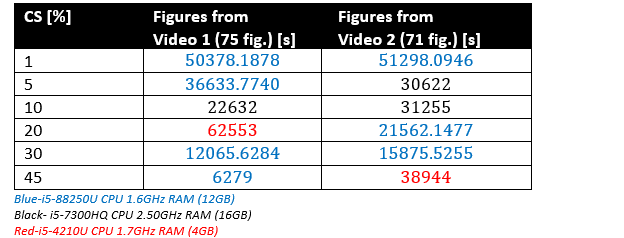}
  \caption{Time spent on Compressive Sensing algorithm of the two videos}
  \label{fig:kompjuteri}
\end{figure}
\subsection*{Experiment II}
\begin{figure}[h!]
  \includegraphics[width=\linewidth]{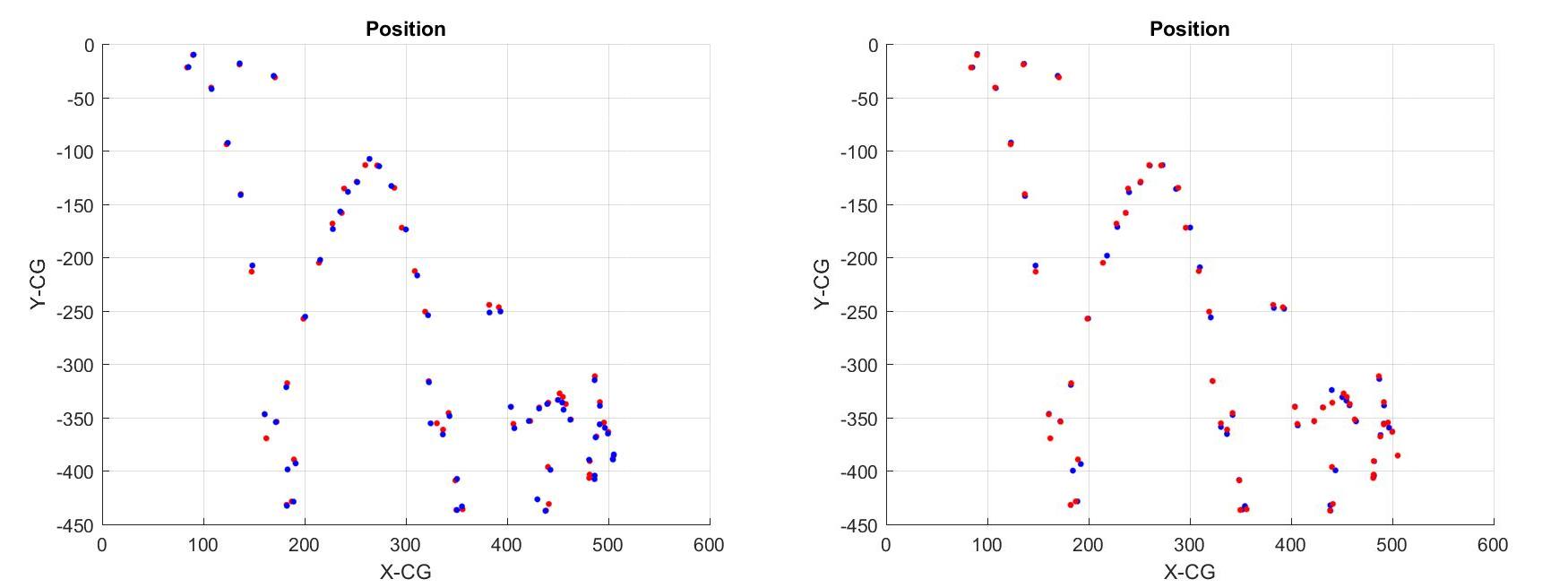}
  \caption{Trajectory of the ball reconstucted from video of 1\%(left) and 5\%(right) of original pixels}
  \label{fig:Video2-rez1}

  \includegraphics[width=\linewidth]{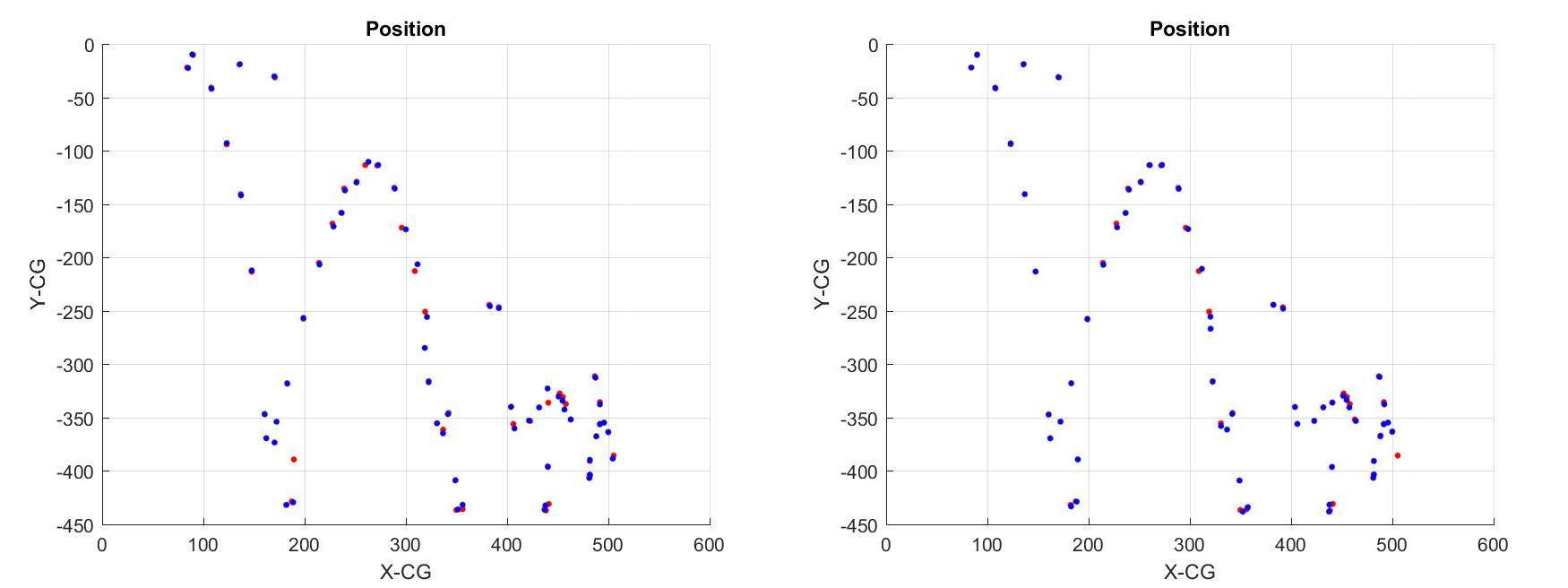}
  \caption{Trajectory of the ball reconstucted from video of 10\%(left) and 20\%(right) of original pixels}
  \label{fig:Video2-rez2}

  \includegraphics[width=\linewidth]{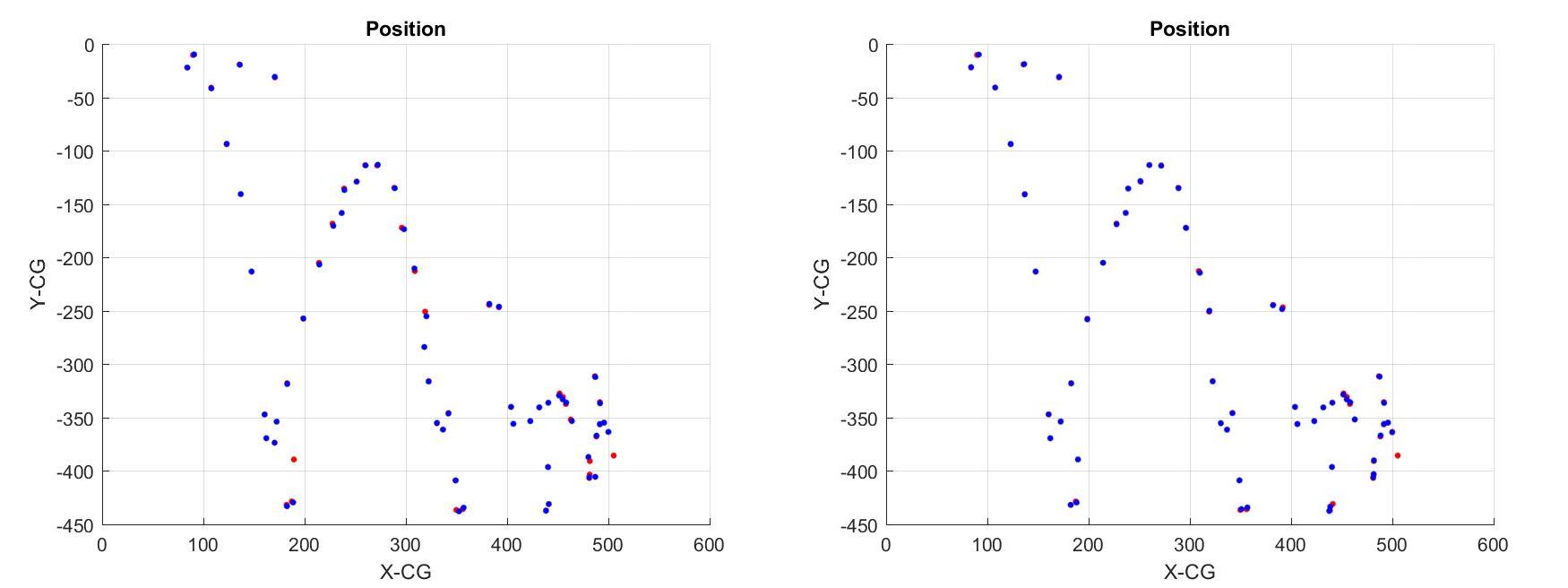}
  \caption{Trajectory of the ball reconstucted from video of 30\%(left) and 45\%(right) of original pixels}
  \label{fig:Video2-rez3}
\end{figure}

\begin{figure}[!htb]
\minipage{0.235\textwidth}
  \includegraphics[width=\linewidth]{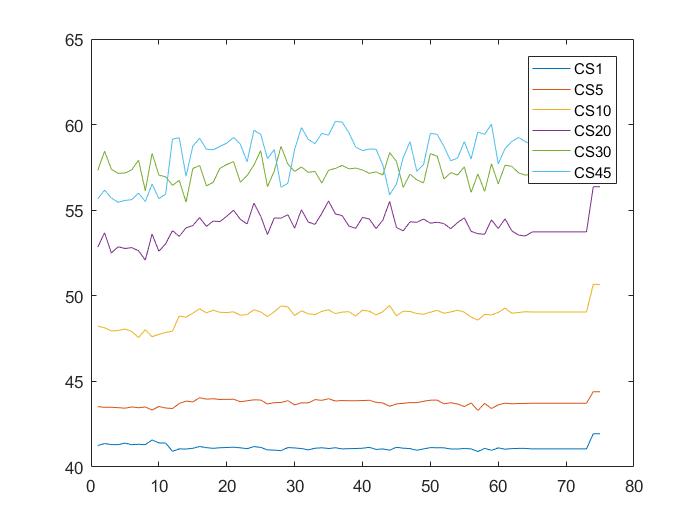}
  \caption{PSNR of the first video with different percentage of pixels conserved}\label{fig:Video1-psnr}
\endminipage\hfill
\minipage{0.235\textwidth}
  \includegraphics[width=\linewidth]{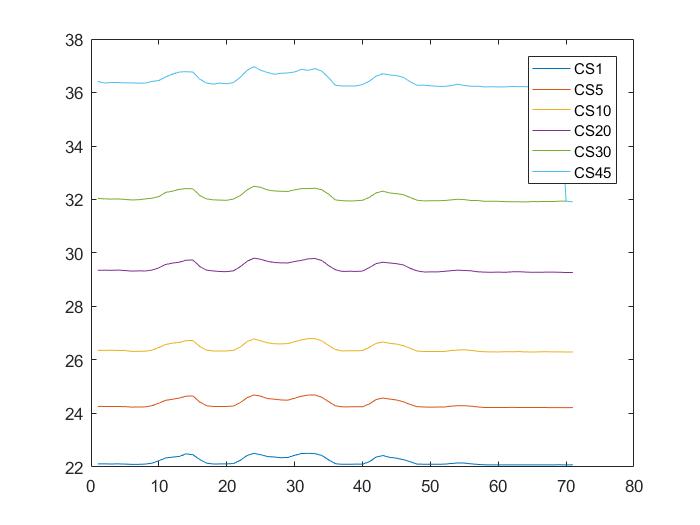}
  \caption{PSNR of the second video with different percentage of pixels conserved}\label{fig:Video2-psnr}
\endminipage
\end{figure}

The second experiment is conducted using a much more complicated video than the previous one. The motion of the ball is more complex, the background is diverse and the object has a shadow (figure \ref{fig:Video2-ORG}).\\
Duration of the video is 2.3667 seconds, size of the video frame is 512x512 and frame rate is 30 frames per second.\\
In figures \ref{fig:Video2-rez1} to \ref{fig:Video2-rez3} we can see that with the decrease of percentage of pixels retained,the deviation is bigger. With only 45\% of retained pixels complete match is achieved while with 1\% of retained pixels the object can be tracked with decent precision, but the trace is shifted from the original.\\
Experiments were done using three different computers with different specifications. Table provided in fig \ref{fig:kompjuteri} shows the elapsed time of the compressive sensing algorithm for different percentages.\\
In addition we are presenting PSNR (Peak signal-to-noise ratio) results in figures \ref{fig:Video1-psnr} and \ref{fig:Video2-psnr}. As the figures show, we will get a high PSNR when compressive sensing is introduced to the first video. That means that the video will retain its quality even when 1\% of pixels are sensed. That is expected when taken the ‘simplicity’ of the first video into consideration. But when it comes to the second experiment the results are quite different. Given the complexity of the second video, when 1\% of pixels are used more errors are introduced. The first video has a much higher PSNR than the second video hence the quality of the first video is better. When 45\% of pixels are contained, the first video has a PSNR of about 80dB and the second video has a PSNR of about 30dB.
\section{Conclusion}
In this paper, we preformed compressive sensing and object tracking on two videos . The second video was less sparse than the first one. The first video has less details, a blank background and smoother movement. All that led to the conclusion that even with 1\% of pixels retained the video closely resembled the original. And the object tracking was still pretty accurate. Due to the complexity of the second video, the results were not as satisfactory as they were with the first one.  Even with 10\% of pixels conserved a clear decline in quality can be seen. This results in shifted reconstructed path, of the targeted object, when compared to the original. Nevertheless, if this algorithm is used when rigorous accuracy isn’t the main goal, the results are still adequate.\\

\end{document}